\documentstyle[lrec2000]{article}

\title{PRIME: A System for Multi-lingual Patent Retrieval}

\name{Shigeto Higuchi$^{\dagger}$, Masatoshi Fukui$^{\dagger}$, 
\large\bf Atsushi Fujii$^{\dagger\dagger,\dagger\dagger\dagger}$, and Tetsuya
Ishikawa$^{\dagger\dagger}$}

\address{$^{\dagger}$PATOLIS Corporation \\
2-4-29 Shiohama Koto-ku, 135-0043, Japan \\
$^{\dagger\dagger}$University of Library and Information Science \\
1-2 Kasuga Tsukuba, 305-8550, Japan \\
$^{\dagger\dagger\dagger}$CREST, Japan Science and Technology Corporation \\
fujii@ulis.ac.jp}

\abstract{Given the growing number of patents filed in multiple
countries, users are interested in retrieving patents across
languages.  We propose a multi-lingual patent retrieval system, which
translates a user query into the target language, searches a
multilingual database for patents relevant to the query, and improves
the browsing efficiency by way of machine translation and
clustering. Our system also extracts new translations from patent
families consisting of comparable patents, to enhance the translation
dictionary.}


\newcommand{\eq}[1]{(\ref{#1})}

\input{psfig.tex}

\begin{document}

\maketitleabstract

\section{Introduction}
\label{sec:introduction}

Given the growing number of patents filed in multiple countries, it is
feasible that users are interested in retrieving patent information
across languages. However, many users find it difficult to perform
patent retrieval (i.e., formulating queries, searching databases for
relevant patents, and browsing retrieved patents) in foreign
languages.

To counter this problem, cross-language information retrieval (CLIR),
where queries in one language are submitted to retrieve documents in
another language, can be an effective solution.  CLIR has of late
become one of the major topics within the information retrieval and
natural language processing communities. In fact, a number of
methods/systems for CLIR have been proposed.

Since by definition queries and documents are in different languages,
queries and documents need to be standardized into a common
representation, so that monolingual retrieval techniques can be
applied. From this point of view, existing CLIR methods are classified
into the following three fundamental categories.

The first method translates queries into the document
language~\cite{ballesteros:sigir-98,fujii:chum-x,nie:sigir-99}, and
the second method translates documents into the query
language~\cite{mccarley:acl-99,oard:amta-98}. The third method
projects both queries and documents into a language-independent space
by way of thesaurus classes~\cite{gonzalo:chum-98,salton:jasis-70} and
latent semantic indexing~\cite{carbonell:ijcai-97,littman:clir-98}.

Among those above methods, the first one (i.e., query translation
method) is preferable in terms of implementation cost, because this
approach can simply be combined with existing monolingual retrieval
systems.

Following a query translation
method~\cite{fujii:emnlp-vlc-99,fujii:chum-x}, we previously proposed
a Japanese/English cross-language patent retrieval
system~\cite{fukui:sigir-ws-pr-2000}, where users submit queries in
either Japanese or English to retrieve patents in the other language.
In either case, the target database is monolingual.

However, since users are not always sure as to which language database
contains patents relevant to their information need, it is effective
to retrieve patents in multiple languages {\em simultaneously\/}.
This process, which we shall call ``multi-lingual information
retrieval (MLIR)'', is an extension of CLIR. In this paper, we propose
a Japanese/English multi-lingual patent retrieval system called
``PRIME'' (Patent Retrieval In Multi-lingual Environment),

The design of our system is based on that for technical
documents~\cite{fujii:ntcir-2-2001}, which combines query translation,
document retrieval, document translation and clustering modules
(Section~\ref{sec:system}).

Additionally, in this paper we newly introduce a module for enhancing
a dictionary used for the query translation module. For this purpose,
we propose a method to extract Japanese/English translations from
patent families consisting of comparable patents filed in Japan and
the United States (Section~\ref{sec:extraction}).

\section{System Description}
\label{sec:system}

\subsection{Overview}
\label{subsec:system_overview}

\begin{figure}[htbp]
  \begin{center}
    \leavevmode
    \psfig{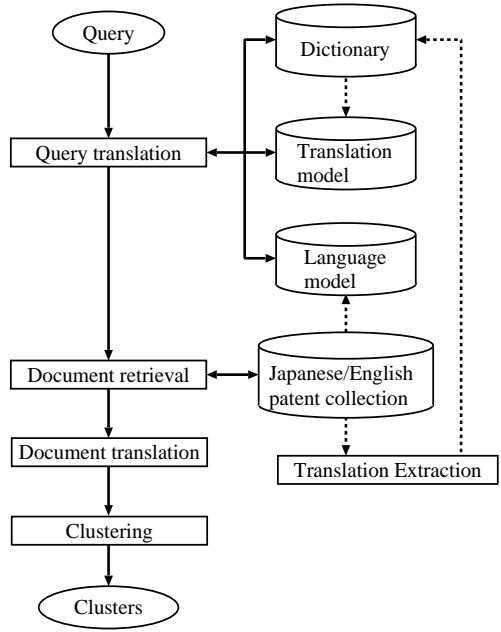}
  \end{center}
  \caption{The design of PRIME: our multi-lingual patent retrieval system
  (dashed arrows denote the off-line process).}
  \label{fig:system}
\end{figure}

Figure~\ref{fig:system} depicts the overall design of PRIME, which
retrieves documents in response to user queries in either Japanese or
English. However, unlike the case of CLIR, retrieved documents can
potentially be in either a combination of Japanese and English or
either of the languages individually. We briefly explain the entire
on-line process based on this figure.

First, a user query is translated into the foreign language (i.e.,
either Japanese or English) by way of a query translation module.

Second, a document retrieval module uses both the source (user) and
translated queries to search a Japanese/English bilingual patent
collection for relevant documents.

In real world usage, Japanese and English patents are not comparable
in the collection (this is the major reason why cross/multi-lingual
retrieval is needed). However, for the purpose of research and
development, we currently target a comparable collection.

To put it more precisely, the collection contains approximately
1,750,000 pairs of Japanese abstracts and their English translations,
which were provided on PAJ (Patent Abstract of Japan) CD-ROMs in
1995-1999\footnote{Copyright by Japan Patent Office.}.

Third, among retrieved documents, only those that are in the foreign
language are translated into the user language through a document
translation module.

In principle, we need only above three modules to realize
multi-lingual patent retrieval in the sense that users can
retrieve/browse foreign documents through their native
language. However, to improve the browsing efficiency, a clustering
module finally divides retrieved documents into a specific number of
groups.

Additionally, in the off-line process, a translation extraction module
identifies Japanese/English translations in the database, to enhance
the query translation module.

\subsection{Query Translation}
\label{subsec:query_translation}

The query translation module is based on the method proposed by Fujii
and Ishikawa~\shortcite{fujii:emnlp-vlc-99,fujii:chum-x}, which has
been applied to Japanese/English CLIR for the NTCIR collection
consisting of technical abstracts~\cite{kando:sigir-99}.

This method translates words and phrases (compound words) in a given
query, maintaining the word order in the source language.  A
preliminary study showed that approximately 95\% of compound technical
terms defined in a bilingual dictionary~\cite{ferber:89} maintain the
same word order in both Japanese and English.

Then, the Nova dictionary\footnote{Developed by NOVA,
Inc. http://www.nova.co.jp/} is used to derive possible word/phrase
translations, and a probabilistic method is used to resolve
translation ambiguity.

The Nova dictionary includes approximately one million
Japanese-English translations related to 19 technical fields as listed
below:
\begin{quote}
  aeronautics, biotechnology, business, chemistry, computers,
  construction, defense, ecology, electricity, energy, finance, law,
  mathematics, mechanics, medicine, metals, oceanography, plants,
  trade.
\end{quote}

In addition, for words unlisted in the Nova dictionary,
transliteration is performed to identify phonetic equivalents in the
target language. Since Japanese often represents loanwords (i.e.,
technical terms and proper nouns imported from foreign languages)
using its special phonetic alphabet (or phonogram) called ``{\it
katakana}'', with which new words can be spelled out, transliteration
is effective to improve the translation quality.

We represent the user query and one translation candidate in the
document language by $U$ and $D$, respectively.  From the viewpoint of
probability theory, our task here is to select $D$'s with greater
probability, $P(D|U)$, which can be transformed as in
Equation~\eq{eq:query_translation} through the Bayesian theorem.
\begin{equation}
  \label{eq:query_translation}
  P(D|U) = \frac{\textstyle P(U|D)\cdot P(D)}{\textstyle P(U)}
\end{equation}
In practice, $P(U)$ can be omitted because this factor is a constant
with respect to the given query, and thus does not affect the relative
probability for different translation candidates.

$P(D)$ is estimated by a word-based bi-gram language model produced
from the target collection. $P(U|D)$ is estimated based on the word
frequency obtained from the Nova dictionary.  Those two factors are
commonly termed language and translation models, respectively (see
Figure~\ref{fig:system}).

\subsection{Document Retrieval}
\label{subsec:retrieval}

The retrieval module is based on an existing probabilistic retrieval
method~\cite{robertson:sigir-94}, which computes the relevance score
between the translated query and each document in the collection. The
relevance score for document $i$ is computed based on
Equation~\eq{eq:okapi}.
\begin{equation}
  \label{eq:okapi}
  \sum_{t} \left(\frac{\textstyle TF_{t,i}}{\textstyle
    \frac{\textstyle DL_{i}}{\textstyle avglen} +
    TF_{t,i}}\cdot\log\frac{\textstyle N}{\textstyle DF_{t}}\right)
\end{equation}
Here, $TF_{t,i}$ denotes the frequency that term $t$ appears in
document $i$. $DF_{t}$ and $N$ denote the number of documents
containing term $t$ and the total number of documents in the
collection. $DL_{i}$ denotes the length of document $i$ (i.e., the
number of characters contained in $i$), and $avglen$ denotes the
average length of documents in the collection.

For both Japanese and English collections, we use content words
extracted from documents as terms, and perform a word-based indexing.
For the Japanese collection, we use the ChaSen morphological
analyzer~\cite{matsumoto:chasen-99} to extract content words. However,
for the English collection, we extract content words based on
parts-of-speech as defined in WordNet~\cite{fellbaum:wordnet-98}.

\subsection{Document Translation}
\label{subsec:document_translation}

The document translation module consists of the the Transer
Japanese/English MT system, which uses the same dictionary used for
the query translation module.

In practice, since machine translation is computationally expensive
and degrades the time efficiency, we perform machine translation on a
phrase-by-phrase basis.  In brief, phrases are sequences of content
words in documents, for which we developed rules to generate phrases
based on the part-of-speech information. This method is practical
because even a word/phrase-based translation can potentially improve
on the efficiency for users to find relevant foreign documents from
the whole retrieval result~\cite{oard:ipm-99}.

\subsection{Clustering}
\label{subsec:clustering}

For the purpose of clustering retrieved documents, we use the
Hierarchical Bayesian Clustering (HBC) method~\cite{iwayama:ijcai-95},
which merges similar items (i.e., documents in our case) in a
bottom-up manner, until all the items are merged into a single
cluster. Thus, a specific number of clusters can be obtained by
splitting the resultant hierarchy at a predetermined level.

The HBC method also determines the most representative item (centroid)
for each cluster. Thus, we can enhance the browsing efficiency by
presenting only those centroids to users.

The similarity between documents is computed based on feature vectors
that characterize each document. In our case, vectors for each
document consist of frequencies of content words appearing in the
document. We extract content words from documents as performed in
word-based indexing (see Section~\ref{subsec:retrieval}).

Given the clustering module, the system can facilitate an interactive
retrieval. To put it more precisely, through the interface, users can
discard irrelevant clusters determined by browsing representative
documents, and re-cluster the remaining documents. By performing this
process recursively, relevant documents are eventually remained.

\section{Extracting Translations Using Patent Families}
\label{sec:extraction}

\subsection{Overview}
\label{subsec:extraction_overview}

Since patents are usually associated with new words, it is crucial to
translate out-of-dictionary words. The transliteration method used in
the query translation module is one solution for this problem (see
Section~\ref{subsec:query_translation}).

On the other hand, it is also effective to update the translation
dictionary.  For this purpose, a number of methods to extract
translations from bilingual (parallel/comparable)
corpora~\cite{smadja:cl-96,yamamoto:coling-2000} are
applicable. However, it is considerably expensive to obtain bilingual
corpora with sufficient volume of alignment information.

To resolve this problem, we use patent families, which are patent sets
filed for the same/related contents in multiple countries, as
comparable corpora. Thus, patents contained in the same family are not
necessarily parallel, but quite comparable.

Among a number of ways to apply for patents in multiple countries, we
focus solely on patents claiming priority under the Paris Convention,
because we can easily identify patent families by the identification
number assigned to each patent.

In addition, the number of patent families is still increasing. Thus,
we can easily update a large-scale bilingual comparable corpus based
on patent families. To the best of our knowledge no research has
utilized patent families for extracting translations.

\subsection{Methodology}
\label{subsec:extraction_method}

Since patents are structured with a number of fields (e.g., titles,
abstracts, and claims), our method first identifies corresponding
fragments based on the document structure, to improve the extraction
accuracy.

However, structures of paired patents are not always the same. For
example, the number of fields claimed in a single patent family often
varies depending on the language. Thus, we use only the title and
abstract fields, which usually parallel in Japanese and English
patents. In other words, unlike the case of most existing extraction
methods, our method does not need sentence-aligned corpora.

We use the ChaSen morphological analyzer~\cite{matsumoto:chasen-99}
and Brill tagger~\cite{brill:cl-95} to extract content words from
Japanese and English fragments, respectively. In addition, we combine
more than one word into phrases, for which we developed rules to
generate phrases based on the part-of-speech information.

We then compute the association score for all the possible
combinations of Japanese/English phrases co-occurring in the same
fragment, and select those with greater score as the final
translations.  For this purpose, we use the weighted Dice
coefficient~\cite{yamamoto:coling-2000} as shown in
Equation~\eq{eq:wdice}.
\begin{equation}
  \label{eq:wdice}
  score(W_{j}, W_{e}) = \log F_{je}\cdot\frac{\textstyle 2
  F_{je}}{\textstyle F_{j} + F_{e}}
\end{equation}
Here, $W_{j}$ and $W_{e}$ are Japanese and English phrases,
respectively. $F_{j}$ and $F_{e}$ denote the frequency that $W_{j}$
and $W_{e}$ appear in the entire corpus, respectively.  $F_{je}$
denotes the frequency that $W_{j}$ and $W_{e}$ co-occur in the same
fragment. The logarithm factor is effective to discard infrequent
co-occurrences, which usually decrease the extraction accuracy.

\subsection{Experimentation}
\label{subsec:extraction_experimentation}

A preliminary study showed that out of approximately 1,750,000 patents
filed in Japan (1995-1999), approximately 32,000 patents were paired
with those filed in the United States as patent families. Thus, in
practice we obtained a bilingual comparable corpus consisting of
32,000 Japanese/English pairs.  From this corpus, our method extracted
1,234,347 phrase-based translations, which were judged it correct or
incorrect.

However, we selected translations association whose score was above
1.5, and manually judged their correctness, because a) the judgement
can be considerably expensive for the entire translations, and b)
translations with small association scores are usually incorrect.  The
total number of selected translations was 37,669.

We then evaluated the accuracy of our extraction method. The accuracy
is the ratio between the number of correct translations, and the
number of cases where the association score of the translation is
above a specific threshold.  By raising the value of the threshold,
the accuracy also increased, while the number of extracted
translations decreased, as shown in Table~\ref{tab:extraction}.
According to this table, we could achieve a high accuracy by limiting
the number of translations extracted.

We spent only four man-days in judging the 37,669 translations and
identifying 5,879 correct translations.  In other words, our method
facilitated to produce bilingual lexicons semi-automatically with a
trivial cost.

\begin{table}[htbp]
  \begin{center}
    \caption{Accuracy for translation extraction.}
    \medskip
    \leavevmode
    \small
    \tabcolsep=2pt
    \begin{tabular}{lrrrrr} \hline\hline
      Threshold for Score & 1.5 & 2.0 & 3.0 & 4.0 & 5.0 \\ \hline
      \# of Translations & 37,669 & 24,869 & 4,419 & 962 & 356 \\
      \# of Correct Translations & 5,879 & 4,129 & 1,399 & 564 & 240 \\
      Accuracy (\%) & 15.6 & 16.6 & 31.7 & 58.6 & 67.4 \\
      \hline
    \end{tabular}
    \label{tab:extraction}
  \end{center}
\end{table}

\section{Conclusion}
\label{sec:conclusion}

In this paper, we proposed a multi-lingual system for Japanese/English
patent retrieval. For this purpose, we used a query translation method
explored in cross-language information retrieval (CLIR).

However, unlike the case of CLIR, our system retrieves bilingual
patents simultaneously in response to a monolingual query.  Our system
also summarizes retrieved patents by way of machine translation and
clustering to improve the browsing efficiency.

In addition, our system includes an extraction module which produces
new translations from patent families consisting of comparable
patents, and updates the translation dictionary.

Future work would include improving existing modules in our system,
and the application of our framework to other languages.

\section*{Acknowledgments}

The authors would like to thank NOVA, Inc. for their support with the
Nova dictionary and Transer system, and Makoto Iwayama for his support
with the HBC software.

\small
\bibliographystyle{acl}

\begin{thebibliography}{}

\bibitem[\protect\citename{Ballesteros and Croft}1998]{ballesteros:sigir-98}
Lisa Ballesteros and W.~Bruce Croft.
\newblock 1998.
\newblock Resolving ambiguity for cross-language retrieval.
\newblock In {\em Proceedings of the 21st Annual International ACM SIGIR
  Conference on Research and Development in Information Retrieval}, pages
  64--71.

\bibitem[\protect\citename{Brill}1995]{brill:cl-95}
Eric Brill.
\newblock 1995.
\newblock Transformation-based error-driven learning and natural language
  processing: A case study in part-of-speech tagging.
\newblock {\em Computational Linguistics}, 21(4):543--565.

\bibitem[\protect\citename{Carbonell \bgroup et al.\egroup
  }1997]{carbonell:ijcai-97}
Jaime~G. Carbonell, Yiming Yang, Robert~E. Frederking, Ralf~D. Brown, Yibing
  Geng, and Danny Lee.
\newblock 1997.
\newblock Translingual information retrieval: A comparative evaluation.
\newblock In {\em Proceedings of the 15th International Joint Conference on
  Artificial Intelligence}, pages 708--714.

\bibitem[\protect\citename{Fellbaum}1998]{fellbaum:wordnet-98}
Christiane Fellbaum, editor.
\newblock 1998.
\newblock {\em {WordNet}: An Electronic Lexical Database}.
\newblock MIT Press.

\bibitem[\protect\citename{Ferber}1989]{ferber:89}
Gene Ferber.
\newblock 1989.
\newblock {\em {English-Japanese}, {Japanese-English} Dictionary of Computer
  and Data-Processing Terms}.
\newblock MIT Press.

\bibitem[\protect\citename{Fujii and Ishikawa}1999]{fujii:emnlp-vlc-99}
Atsushi Fujii and Tetsuya Ishikawa.
\newblock 1999.
\newblock Cross-language information retrieval for technical documents.
\newblock In {\em Proceedings of the Joint ACL SIGDAT Conference on Empirical
  Methods in Natural Language Processing and Very Large Corpora}, pages 29--37.

\bibitem[\protect\citename{Fujii and Ishikawa}2001]{fujii:ntcir-2-2001}
Atsushi Fujii and Tetsuya Ishikawa.
\newblock 2001.
\newblock Evaluating multi-lingual information retrieval and clustering at
  {ULIS}.
\newblock In {\em Proceedings of the 2nd NTCIR Workshop Meeting on Evaluation
  of Chinese \& Japanese Text Retrieval and Text Summarization}.

\bibitem[\protect\citename{Fujii and Ishikawa}To appear]{fujii:chum-x}
Atsushi Fujii and Tetsuya Ishikawa.
\newblock (To appear).
\newblock {Japanese/English} cross-language information retrieval: Exploration
  of query translation and transliteration.
\newblock {\em Computers and the Humanities}.

\bibitem[\protect\citename{Fukui \bgroup et al.\egroup
  }2000]{fukui:sigir-ws-pr-2000}
Masatoshi Fukui, Shigeto Higuchi, Youichi Nakatani, Masao Tanaka, Atsushi
  Fujii, and Tetsuya Ishikawa.
\newblock 2000.
\newblock Applying a hybrid query translation method to {Japanese/English}
  cross-language patent retrieval.
\newblock In {\em ACM SIGIR Workshop on Patent Retrieval}.

\bibitem[\protect\citename{Gonzalo \bgroup et al.\egroup
  }1998]{gonzalo:chum-98}
Julio Gonzalo, Felisa Verdejo, Carol Peters, and Nicoletta Calzolari.
\newblock 1998.
\newblock Applying {EuroWordNet} to cross-language text retrieval.
\newblock {\em Computers and the Humanities}, 32:185--207.

\bibitem[\protect\citename{Iwayama and Tokunaga}1995]{iwayama:ijcai-95}
Makoto Iwayama and Takenobu Tokunaga.
\newblock 1995.
\newblock Hierarchical {Bayesian} clustering for automatic text classification.
\newblock In {\em Proceedings of the 14th International Joint Conference on
  Artificial Intelligence}, pages 1322--1327.

\bibitem[\protect\citename{Kando \bgroup et al.\egroup }1999]{kando:sigir-99}
Noriko Kando, Kazuko Kuriyama, and Toshihiko Nozue.
\newblock 1999.
\newblock {NACSIS} test collection workshop ({NTCIR-1}).
\newblock In {\em Proceedings of the 22nd Annual International ACM SIGIR
  Conference on Research and Development in Information Retrieval}, pages
  299--300.

\bibitem[\protect\citename{Littman \bgroup et al.\egroup
  }1998]{littman:clir-98}
Michael~L. Littman, Susan~T. Dumais, and Thomas~K. Landauer.
\newblock 1998.
\newblock Automatic cross-language information retrieval using latent semantic
  indexing.
\newblock In Gregory Grefenstette, editor, {\em Cross-Language Information
  Retrieval}, chapter~5, pages 51--62. Kluwer Academic Publishers.

\bibitem[\protect\citename{Matsumoto \bgroup et al.\egroup
  }1999]{matsumoto:chasen-99}
Yuji Matsumoto, Akira Kitauchi, Tatsuo Yamashita, Yoshitaka Hirano, Hiroshi
  Matsuda, and Masayuki Asahara.
\newblock 1999.
\newblock {Japanese} morphological analysis system {ChaSen} version 2.0 manual
  2nd edition.
\newblock Technical Report NAIST-IS-TR99009, NAIST.

\bibitem[\protect\citename{McCarley}1999]{mccarley:acl-99}
J.~Scott McCarley.
\newblock 1999.
\newblock Should we translate the documents or the queries in cross-language
  information retrieval?
\newblock In {\em Proceedings of the 37th Annual Meeting of the Association for
  Computational Linguistics}, pages 208--214.

\bibitem[\protect\citename{Nie \bgroup et al.\egroup }1999]{nie:sigir-99}
Jian-Yun Nie, Michel Simard, Pierre Isabelle, and Richard Durand.
\newblock 1999.
\newblock Cross-language information retrieval based on parallel texts and
  automatic mining of parallel texts from the {Web}.
\newblock In {\em Proceedings of the 22nd Annual International ACM SIGIR
  Conference on Research and Development in Information Retrieval}, pages
  74--81.

\bibitem[\protect\citename{Oard and Resnik}1999]{oard:ipm-99}
Douglas~W. Oard and Philip Resnik.
\newblock 1999.
\newblock Support for interactive document selection in cross-language
  information retrieval.
\newblock {\em Information Processing \& Management}, 35(3):363--379.

\bibitem[\protect\citename{Oard}1998]{oard:amta-98}
Douglas~W. Oard.
\newblock 1998.
\newblock A comparative study of query and document translation for
  cross-language information retrieval.
\newblock In {\em Proceedings of the 3rd Conference of the Association for
  Machine Translation in the Americas}, pages 472--483.

\bibitem[\protect\citename{Robertson and Walker}1994]{robertson:sigir-94}
S.~E. Robertson and S.~Walker.
\newblock 1994.
\newblock Some simple effective approximations to the 2-poisson model for
  probabilistic weighted retrieval.
\newblock In {\em Proceedings of the 17th Annual International ACM SIGIR
  Conference on Research and Development in Information Retrieval}, pages
  232--241.

\bibitem[\protect\citename{Salton}1970]{salton:jasis-70}
Gerard Salton.
\newblock 1970.
\newblock Automatic processing of foreign language documents.
\newblock {\em Journal of the American Society for Information Science},
  21(3):187--194.

\bibitem[\protect\citename{Smadja \bgroup et al.\egroup }1996]{smadja:cl-96}
Frank Smadja, Kathleen~R. McKeown, and Vasileios Hatzivassiloglou.
\newblock 1996.
\newblock Translating collocations for bilingual lexicons: A statistical
  approach.
\newblock {\em Computational Linguistics}, 22(1):1--38.

\bibitem[\protect\citename{Yamamoto and Matsumoto}2000]{yamamoto:coling-2000}
Kaoru Yamamoto and Yuji Matsumoto.
\newblock 2000.
\newblock Acquisition of phrase-level bilingual correspondence using dependency
  structure.
\newblock In {\em Proceedings of the 18th International Conference on
  Computational Linguistics}, pages 933--939.

\end{thebibliography}

\end{document}